\documentclass[10pt]{article}
\usepackage{ml98}

\title{An Investigation of Transformation-Based Learning in Discourse} 
 
\author{ {\bf Ken Samuel} \\  
CIS Department \\
University of Delaware \\ 
Newark, Delaware 19716 USA \\ 
samuel@cis.udel.edu \\
\And 
{\bf Sandra Carberry}  \\ 
CIS Department \\
University of Delaware \\ 
Newark, Delaware 19716 USA \\ 
carberry@cis.udel.edu \\
\And 
{\bf K. Vijay-Shanker}   \\ 
CIS Department \\
University of Delaware \\ 
Newark, Delaware 19716 USA \\ 
vijay@cis.udel.edu \\
}

\input{psfig}

\begin{document} 
 
\maketitle 

\begin{abstract} 

This paper presents results from the first attempt to apply
Transformation-Based Learning to a discourse-level Natural Language
Processing task. To address two limitations of the standard algorithm,
we developed a Monte Carlo version of Transformation-Based Learning to
make the method tractable for a wider range of problems without
degradation in accuracy, and we devised a committee method for
assigning confidence measures to tags produced by Transformation-Based
Learning. The paper describes these advances, presents experimental
evidence that Transformation-Based Learning is as effective as
alternative approaches (such as Decision Trees and N-Grams) for a
discourse task called Dialogue Act Tagging, and argues that
Transformation-Based Learning has desirable features that make it
particularly appealing for the Dialogue Act Tagging task.

\end{abstract} 
 
\section{INTRODUCTION} 
 
Transformation-Based Learning is a relatively new machine learning
method, which has been as effective as any other approach on the
Part-of-Speech Tagging problem\footnote{The goal of this Natural
Language Processing task is to label words with the proper part of
speech tags, such as Noun and Verb.}~(Brill, 1995a). We are utilizing
Transformation-Based Learning for another important language task
called Dialogue Act Tagging, in which the goal is to label each
utterance in a conversational dialogue with the proper dialogue act. A
{\em dialogue act} is a concise abstraction of a speaker's intention,
such as SUGGEST or ACCEPT. Recognizing dialogue acts is critical for
discourse-level understanding and can also be useful for other
applications, such as resolving ambiguity in speech recognition. But
computing dialogue acts is a challenging task, because often a
dialogue act cannot be directly inferred from a literal reading of an
utterance. Figure~\ref{ex-das} presents a hypothetical dialogue that
has been labeled with dialogue acts.

\begin{figure*}[ht]
\centering
\begin{tabular}{|cclc|}
\multicolumn{1}{c}{\#} & Speaker & \multicolumn{1}{c}{Utterance} &
\multicolumn{1}{c}{Dialogue Act} \\
\hline
1 & John & Hello.                                          & GREET \\
2 & John & I'd like to meet with you on Tuesday at 2:00.   & SUGGEST \\
3 & Mary & That's no good for me,                          & REJECT \\
4 & Mary & but I'm free at 3:00.                           & SUGGEST \\
5 & John & That sounds fine to me.                         & ACCEPT \\
6 & John & I'll see you then.                              & BYE \\
\hline
\end{tabular}
\caption{A sample dialogue}
\label{ex-das}
\end{figure*}

Our research efforts led us to address some limitations of
Transformation-Based Learning. We developed a Monte Carlo version of
the algorithm that overcomes the limitation of Transformation-Based
Learning's dependence on manually-generated {\em rule templates} and
enables Transformation-Based Learning to be applied effectively to a
wider range of tasks. We also devised a technique that uses a
committee of learned models to derive {\em confidence measures}
associated with the dialogue acts assigned to utterances.

We experimentally compared our modified version of
Transformation-Based Learning with C5.0, an implementation of Decision
Trees, and N-Grams, which was previously the best reported method for
Dialogue Act Tagging~(Reithinger and Klesen, 1997). Our system
performs as well as these benchmarks, and we note that
Transformation-Based Learning has several characteristics that make it
particularly appealing for the Dialogue Act Tagging task.

This paper begins with an overview of the Transformation-Based
Learning method, describing the training phase and the application
phase of the algorithm and presenting some of Transformation-Based
Learning's most attractive characteristics for Dialogue Act Tagging.
The following section describes the experimental design used for the
experiments presented in the paper. Then Section 4 presents two
limitations of Transformation-Based Learning, a dependence on rule
templates and a lack of confidence measures, and describes our
solutions for these problems, a Monte Carlo strategy and a committee
method. Next we present an experimental comparison between
Transformation-Based Learning, N-Grams, and Decision Trees, and
conclude with a discussion of this work.

\section{TRANSFORMATION-BASED LEARNING} 
 
Brill~(1995a) developed a symbolic machine learning method called
Transformation-Based Learning. Given a tagged training corpus,
Transformation-Based Learning produces a sequence of rules that
serves as a model of the training data. Then, to derive the
appropriate tags, each rule may be applied, in order, to each instance
in an untagged corpus. For all of the results and examples in this
paper, we are using Transformation-Based Learning on the Dialogue Act
Tagging task, so the instances are utterances and the tags are
dialogue acts. In one experiment, our system produced a learned model
with 213 rules; the first five rules are presented in
Figure~\ref{ex-rules}.

\begin{figure}[ht]
\centering
\begin{tabular}{|c|l|c|}
\hline
   &                                   & New \\
\# & \multicolumn{1}{c|}{Condition(s)} & Dialogue Act \\
\hline
1 & {\em none}                    & SUGGEST\\
\hline
2 & Includes ``see'' and ``you''  & BYE    \\
\hline
3 & Includes ``sounds''           & ACCEPT \\
\hline
4 & Length $<$ 4 words            & GREET  \\
  & Previous tag is {\em none}\footnotemark &        \\
\hline
5 & Includes ``no''               & REJECT \\
  & Previous tag is SUGGEST       &        \\
\hline
\end{tabular}
\caption{Rules produced by Transformation-Based Learning for Dialogue Act Tagging}
\label{ex-rules}
\end{figure}

\footnotetext{This condition is true only for the first utterance of a
dialogue.}

\subsection{THE TRAINING PHASE} 
 
The training phase of TBL, in which the system learns a sequence of
rules based on a tagged training corpus, proceeds in the following
manner:

\smallskip

\begin{ttfamily}

\begin{tabular}{rrrl}
1. & \multicolumn{3}{l}{Label each instance with a dummy tag.} \\
2. & \multicolumn{3}{l}{Until no useful rules are found,} \\
   & a. & \multicolumn{2}{l}{For each incorrect tag} \\
   &    & i. & Generate all rules that  \\
   &    &    & correct the tag. \\
   & b. & \multicolumn{2}{l}{Score each generated rule.} \\
   & c. & \multicolumn{2}{l}{Output the highest scoring rule.} \\
   & d. & \multicolumn{2}{l}{Apply this rule to the corpus.} \\
\end{tabular}

\end{ttfamily}

\smallskip

First, the system initializes the training corpus by labeling each instance
with a dummy tag. Brill~(1995a) suggested using a more complex
initialization step, but we found that this simple strategy is more
effective in practice.\footnote{This is because Transformation-Based
Learning uses an error-driven approach, only generating rules for the
instances that are incorrectly labeled. If every instance is
initialized with a dummy tag, then all of the labels are incorrect,
and so they all contribute to learning. Alternatively, using a more
involved initialization step results in a greater number of correct
tags and, effectively, less training data.} Then the system generates
all of the {\em potential rules} that would make at least one tag in
the training corpus correct, under the restrictions described below.
For each potential rule, its {\em improvement score} is defined to be
the number of correct tags in the training corpus after applying the
rule {\em minus} the number of correct tags in the training corpus
before applying the rule. The potential rule with the highest
improvement score is output as the next rule in the final model and
applied to the entire training corpus. This process repeats (using the
updated tags on the training corpus), producing one rule for each pass
through the training corpus until no rule can be found with an
improvement score that surpasses some predefined threshold. In
practice, threshold values of 1 or 2 appear to be effective.

Since there are potentially an infinite number of rules that could
produce the tags in the training data, it is necessary to
restrict the range of patterns that the system may consider by
providing a set of rule templates, such as:

\begin{tabular}{ll}
IF   & utterance \textbf{u} contains the word(s) \textbf{w} \\
AND  & the tag on the utterance preceding \textbf{u} is \textbf{X} \\
THEN & change \textbf{u}'s tag to \textbf{Y} \\
\end{tabular}

\noindent
This template can be instantiated to produce the last rule in
Figure~\ref{ex-rules} by setting \textbf{w}=``no'', \textbf{X}=SUGGEST, and
\textbf{Y}=REJECT.

For the first rules of the learned model, the emphasis is on getting
as many tags correct as possible with no penalty imposed for changing
an incorrect tag to another incorrect tag. Then for the later rules,
the system must avoid changing any of the tags that are already
correct. Thus, this method tends to produce a sequence of rules that
progresses from general rules to specific rules.

\subsection{THE APPLICATION PHASE}

To see how a rule sequence can be used to label data, consider
applying the rules in Figure~\ref{ex-rules} to the dialogue in
Figure~\ref{ex-das}. The first rule labels every utterance with the
dialogue act SUGGEST. Next, the second rule changes an utterance's tag
to BYE if it contains the words ``see'' and ``you'', which only holds
for utterance \#6. Similarly, the third rule changes utterance \#5's
tag to ACCEPT. Then the fourth rule tags utterance \#1 as GREET, since
its length is 1 and there is no preceding utterance in the dialogue.
And finally, the last rule relabels utterance \#3 as REJECT, since
utterance \#2 is currently tagged SUGGEST, and the word ``no'' is
found in utterance \#3. Although the first five rules label these six
utterances correctly, the remaining 208 rules in the sequence may
continue to adjust the tags on the utterances.

\subsection{ATTRACTIVE CHARACTERISTICS}

For the Dialogue Act Tagging task, we selected Transformation-Based
Learning for several reasons. Brill reported that Transformation-Based
Learning is as good as or better than any other algorithm for the
Part-of-Speech Tagging problem, labeling 97.2\% of the words
correctly. The part-of-speech tag of a word is dependent on the word's
internal features and on the surrounding words; similarly, the
dialogue act of an utterance is dependent on the utterance's internal
features and on the surrounding utterances. This parallel suggests
that Transformation-Based Learning has potential for success on the
Dialogue Act Tagging problem.

Since we currently lack a systematic theory of dialogue acts, another
reason that Transformation-Based Learning is an attractive choice is
that its learned model consists of relatively intuitive rules~(Brill,
1995a), which a human can analyze to determine what the system has
learned and develop a working theory. Also, Transformation-Based
Learning is good at ignoring any potential rules that are irrelevant.
This is because irrelevant rules tend to have a random effect on the
training data, which usually results in low improvement scores, so
these rules are unlikely to be selected for inclusion in the final
model. This is very helpful for Dialogue Act Tagging, since we don't
know what the relevant templates are for this problem. Ramshaw and
Marcus~(1994) experimentally demonstrated Transformation-Based
Learning's robustness with respect to irrelevant rules.

For these reasons, along with others that are presented at the end of
the paper, we believe that Transformation-Based Learning is worthy of
investigation for the Dialogue Act Tagging task.

\section{EXPERIMENTAL DESIGN}

All of the results presented in this paper followed the same
experimental design as the third experiment in Reithinger and
Klesen~(1997). The corpus consisted of appointment-scheduling
face-to-face dialogues in English, which was divided into a training
set with 143 dialogues (2701 utterances) and a disjoint testing set
with 20 dialogues (328 utterances). Each utterance was manually
labeled with one of 18 abstract dialogue acts, such as SUGGEST,
ACCEPT, REJECT, GREET, and BYE. The full list of dialogue
acts is found in Reithinger and Klesen~(1997).

The Transformation-Based Learning experiments presented in this paper
were run on a Sun Ultra 1 machine with 508MB of main memory. Within a
set of experiments, only the specified parameters were varied, but
between sets of experiments many parameters may have been varied, so
it is not possible to draw conclusions across experiment sets.

Our rule templates consist of all possible combinations of a
preselected set of conditions. Some of these conditions are presented
in Figure~\ref{conditions}. Each {\em condition} consists of a feature
and a distance, where the {\em feature} specifies a characteristic of
utterances that might be relevant for the Dialogue Act Tagging task,
and the {\em distance} specifies the relative position (from the
utterance under analysis) of the utterance that the feature should be
applied to.

\begin{figure}[ht]
\centering
\begin{tabular}{ccc}
Feature & & Distance \\
\hline
\hline
length & of the & current utterance \\
\hline
tag & of the & preceding utterance \\
\hline
cue patterns & of the & current utterance \\
\hline
speaker & of the & current utterance \\
\hline
speaker & of the & preceding utterance \\
\end{tabular}
\caption{Some conditions used in our experiments}
\label{conditions}
\end{figure}

In discourse, it is widely acknowledged that some of the short phrases
(and specific words) found in an utterance provide strong clues to
determine the appropriate dialogue act. Several researchers proposed
different {\em cue phrases}, which are phrases that appear frequently
in dialogue and convey useful discourse information, such as ``but'',
``so'', and ``by the way''. Unfortunately, there is no universal
agreement on which phrases should be considered cue phrases, and in a
preliminary experiment using all of the cue phrases proposed in the
literature,\footnote{These lists of cue phrases can be found in
Hirschberg and Litman~(1993) and Knott~(1996).} our system's accuracy
only improved by 1.03\%.

In order to identify the phrases that will be useful for a particular
domain, we need an {\em automatic} method for collecting a set of
phrases that is tuned to that domain. So we are using a statistical
approach to select relevant {\em cue patterns}\footnote{In practice,
the concept of cue patterns tends to be more general than cue phrases,
including many more phrases.} from a training corpus. Assuming that a
phrase is relevant if it co-occurs frequently with a few specific
dialogue acts, we analyze the distribution of dialogue acts for
utterances that include a given phrase, selecting those phrases that
correspond to dialogue act distributions with low entropy. When using
these cue patterns, our system's accuracy rose by 17.63\%. For more
details on this work, see Samuel, Carberry, and Vijay-Shanker~(1998b).

\section{TRANSFORMATION-BASED LEARNING IN DISCOURSE}

\subsection{TWO LIMITATIONS}

Transformation-Based Learning has two serious limitations, which we
will address in this section. First, although Transformation-Based
Learning produces a tag for each instance, it doesn't offer any
measure of confidence in these tags. Alternatively, probabilistic
machine learning approaches generally label an instance with a set of
tags, which are assigned numbers to represent the likelihood that they
are correct. So ``probabilistic methods ... provide a continuous
ranking of alternative analyses rather than just a single output, and
such rankings can productively increase the bandwidth between
components of a modular system.''~(Brill and Mooney, 1997)

\begin{figure*}[ht]
\centerline{\psfig{figure=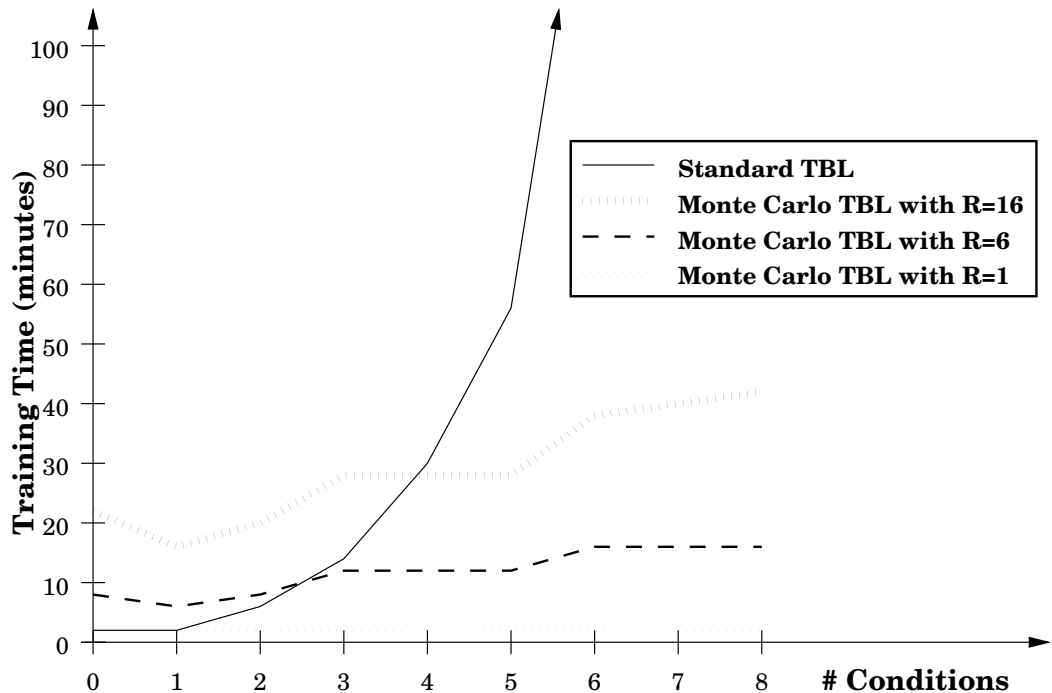}}
\caption{Number of conditions vs. training time}
\label{time-MC-graph}
\end{figure*}

The second limitation of Transformation-Based Learning is that it is
highly dependent on the rule templates, which are manually developed
in advance. Since the omission of any relevant templates would
handicap the system, it is essential that these choices be made
carefully. But in Dialogue Act Tagging, no one knows exactly which
conditions and combinations of conditions are relevant, so it is
preferable to err on the side of caution by constructing an
overly-general set of templates and allowing the system to {\em learn}
which templates are useful. As discussed earlier, Transformation-Based
Learning is capable of discarding irrelevant rules, so this approach
should be effective, in theory.

Unfortunately, this strategy is not tractable, because for each pass
through the training data, for each instance that the system has
tagged incorrectly, {\em every} rule template must be instantiated in
{\em all} possible ways. Suppose that we can postulate \textbf{f}
different features that might be relevant, and we wish to consider
these features for all instances that occur within a distance
\textbf{d} of a given instance. (In other words, we are using a
contextual window of size \textbf{2d+1}.) Then there are
$\mathbf{(2d+1)f}$ conditions and $\mathbf{2^{(2d+1)f}}$ possible
templates, since each condition may either be included or excluded.
Also, suppose that when a feature is applied to an instance, it
produces \textbf{v} distinct values, on average. This results in
$\mathbf{(v+1)^{(2d+1)f}}$ rules per instance, which can be proven by
induction on the number of conditions. Given a training corpus with
\textbf{i} instances, if the algorithm makes \textbf{p} passes through
the training data, then the system must generate and evaluate
$\mathbf{O(ip(v+1)^{(2d+1)f})}$ rules. Some realistic values for these
variables are \textbf{f}=10, \textbf{d}=2 (a contextual window of size
5), \textbf{v}=3, \textbf{i}=3000, and \textbf{p}=100, which generates
around $\mathrm{10^{35}}$ rules. Based on experimental evidence, it
appears that it is necessary to drastically limit the number of
potential rules that the system generates,\footnote{For the
Part-of-Speech Tagging task, Brill used only about 30 simple rule
templates~(Brill, 1995a).} or the memory and time costs are so
exorbitant that the method becomes intractable. But this limitation
would preclude considering all of the features and feature
interactions that might be relevant for Dialogue Act Tagging.

\subsection{A MONTE CARLO VERSION}

\begin{figure*}[ht]
\centerline{\psfig{figure=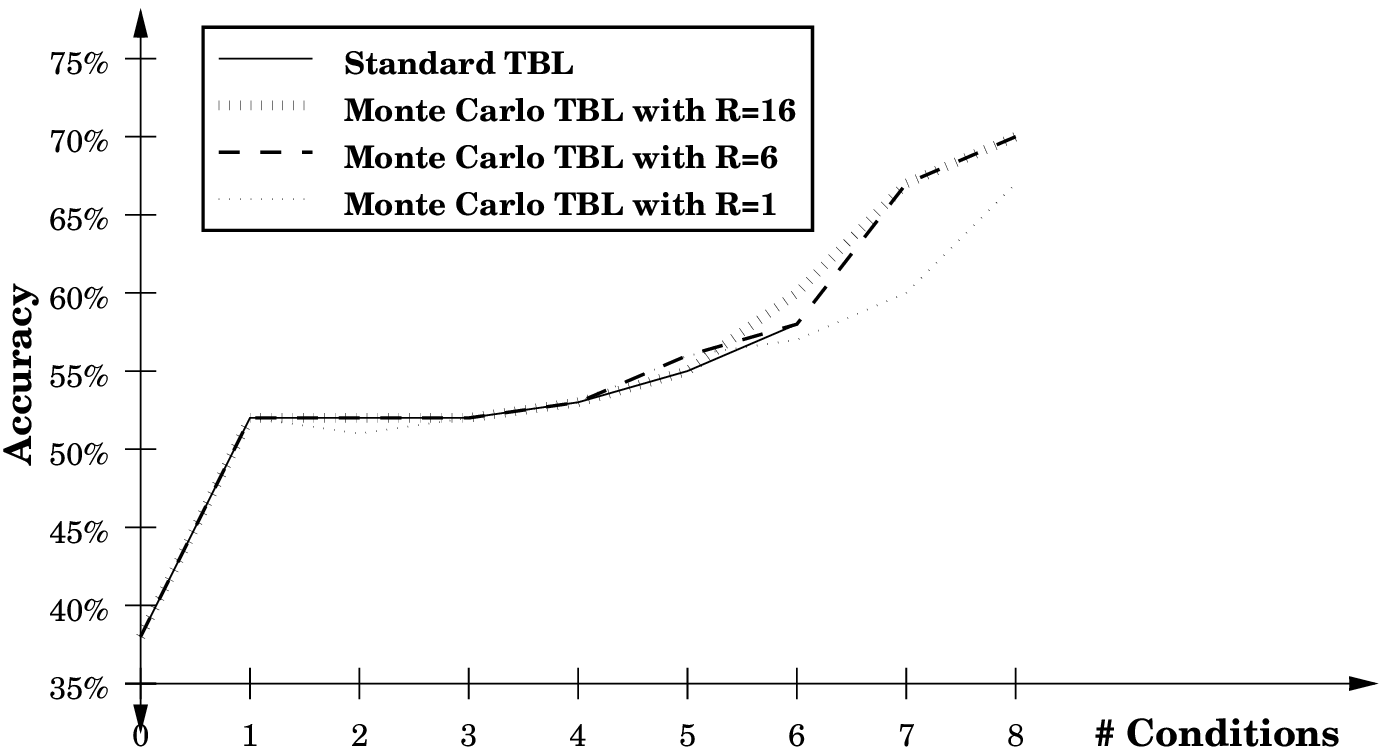}}
\caption{Number of conditions vs. tagging accuracy on unseen data}
\label{accuracy-MC-graph}
\end{figure*}

We developed a Monte Carlo version of Transformation-Based Learning,
so that the system can consider a huge number of templates while still
maintaining tractability. Rather than exhaustively searching through
the space of possible rules, only \textbf{R} of the available template
instantiations are randomly selected for each training instance on
each pass through the training data, where \textbf{R} is some small
integer. With this modification, the total number of rules generated
is only $\mathbf{O(ipR)}$, which no longer explodes with the number of
templates. In fact, the formula doesn't even depend on the number of
features, the contextual window size, or the value of \textbf{v}. But
one would still expect good results, because Transformation-Based
Learning only needs to find the best rules, and the best rules tend to
be effective for a large number of different instances. So the system
has many opportunities to find these rules, and since the algorithm
generally makes many passes through the training data before halting,
if it should select a suboptimal rule, it can use later rules to
compensate. Thus, although random sampling will miss some rules, it is
still highly likely to find an effective sequence of rules.

Our experiments confirm these intuitions, as shown in
Figures~\ref{time-MC-graph} and~\ref{accuracy-MC-graph}. For these
runs, eight conditions were preselected, and for different values of
\textbf{n}, $\mathrm{0 \leq} \mathbf{n} \mathrm{\leq 8}$, the first
\textbf{n} conditions were combined in all possible ways to generate
$\mathbf{2^n}$ templates. Using these templates, we trained, tested,
and compared the standard Transformation-Based Learning method and our
Monte Carlo version of Transformation-Based Learning.

For the standard Transformation-Based Learning method, training time
rises dramatically as the number of conditions increases, as shown in
Figure~\ref{time-MC-graph}.\footnote{The value of \textbf{v} (the
average number of rules generated per instance) varies slightly across
the eight conditions, and so the shape of the curve might vary
depending on the order in which the conditions are presented. But the
critical point is that the training time rises exponentially with the
number of conditions.} In fact, when given seven conditions, the
standard Transformation-Based Learning algorithm could not complete
the training phase, even after running for more than 24 hours. But our
Monte Carlo version of Transformation-Based Learning keeps the
efficiency relatively stable.\footnote{The Monte Carlo version of
Transformation-Based Learning can be slower than the standard method,
because the Monte Carlo version always generates \textbf{R} rules for
each instance, without checking for repetitions. (It would be too
inefficient to prevent the system from generating any rule more than
once.)} The reason for the slight increase in training time as the
number of conditions increases is that, as the system gains access to
a greater number of useful conditions, it's likely to find a greater
number of useful rules, meaning that the training phase makes a
greater number of passes through the training data. Thus, \textbf{p}
increases, and so the training time, \textbf{O(ipR)}, also increases.
But this increase is linear (or less), while standard
Transformation-Based Learning's training time increases exponentially
with the number of conditions. Figure~\ref{time-MC-graph} supports
this analysis.

This improvement in time efficiency would be quite uninteresting if
the performance of the algorithm deteriorated significantly. But, as
Figure~\ref{accuracy-MC-graph} shows, this is not the case. Although
setting \textbf{R} too low (such as \textbf{R}=1 for 7 and 8
conditions) may result in a decrease in accuracy, the lowest possible
setting (\textbf{R}=1) is as accurate as standard Transformation-Based
Learning for 6 conditions (64 templates). For 7 and 8 conditions,
training of the standard Transformation-Based Learning method took too
much time, so those results could not be produced. But, as the curves
for \textbf{R}=6 and \textbf{R}=16 do not differ significantly, it is
reasonable to predict that standard Transformation-Based Learning
would produce similar results as well.\footnote{One might wonder how
the Monte Carlo version of Transformation-Based Learning can ever do
better than the standard Transformation-Based Learning method, which
occurred for the experiments that used five conditions. Because
Transformation-Based Learning is a greedy algorithm, choosing the best
available rule on each pass through the training data, sometimes the
standard Transformation-Based Learning method selects a rule that
locks it into a local maximum, while the Monte Carlo version might
fail to consider this attractive rule and end up producing a better
model.} Therefore, we conclude that our Monte Carlo version of
Transformation-Based Learning (with \textbf{R}=6) works effectively
for more than 250 templates (8 conditions) in only about 15 minutes of
training time.

\subsection{A COMMITTEE METHOD}

We wanted to extend Transformation-Based Learning so that it could
provide some idea of the likelihood that each of its tags are correct.
So we attempted to develop a strategy for assigning confidence
measures to the {\em rules} in the learned model. Then, in the
application phase, a given instance's confidence measure would be a
function of the confidences of the rules that applied to that
instance. Unfortunately, due to the nature of the Transformation-Based
Learning method, this straightforward approach has been unsuccessful,
because the rule sequence does not contain enough information to
derive confidence measures; often, the same pattern of rules applies
to instances that should be marked with high confidence as well as
instances that should be marked with low confidence.

So, for the purpose of computing confidence measures, we adapted two
techniques that were developed for very different tasks. The Boosting
approach has been used to improve accuracy in tagging data~(Freund and
Schapire, 1996), and Committee-Based Sampling utilized a very similar
strategy to minimize the required size of a training corpus~(Dagan and
Engelson, 1995). We applied these methods to compute confidence
measures, by training the system a number of times to produce a few
different but reasonable learned models, which are called {\em
committee members}. Then given new data, each committee member
independently tags the input, and a given tag's confidence is based on
how well the committee members agree on that tag. We are currently
defining the confidence of a given tag to be the number of committee
members that preferred the tag. In the future, we will investigate
confidence formulas that are based on the entropy of the tags selected
by the different committee members.

We considered several ways to develop the committee members, and we
decided to apply the strategy that Freund and Schapire~(1996) used for
Boosting: The first committee member is trained in the standard way,
and then the second committee member pays special attention to those
instances in the training data that the first committee member did not
tag correctly. To do this in Transformation-Based Learning, we adjust
the improvement score formula to weight success on these ``hard''
instances more heavily. (In effect, it is as if we were adding
multiple copies of these instances to the training corpus.) This
process can be repeated to generate more committee members by basing
the score for correctly tagging a training instance on the number of
previous committee members that tagged that instance incorrectly. We
are currently using $\mathbf{2^c}$ as the score for correctly tagging
a given instance that \textbf{c} committee members have mistagged.
This strategy tends to produce committee members that are very
different, as they are focusing on different parts of the training
corpus.

\begin{figure}[ht]
\centering
\begin{tabular}{|c||c|c|}
\hline
Minimum    & Percentage of    & Average \\
Confidence & Instances Tagged & Precision \\
\hline
5 &  45.12\% $\pm$ 1.28\% &  90.09\% $\pm$ 1.51\% \\
4 &  69.79\% $\pm$ 1.60\% &  83.53\% $\pm$ 1.27\% \\
3 &  92.38\% $\pm$ 1.32\% &  76.57\% $\pm$ 0.79\% \\
2 &  99.85\% $\pm$ 0.20\% &  73.56\% $\pm$ 1.10\% \\
1 & 100.00\% $\pm$ 0.00\% &  73.45\% $\pm$ 1.06\% \\
\hline
\end{tabular}
\caption{Testing the committee method on unseen data, varying the
minimum confidence considered}
\label{confidences}
\end{figure}

As a preliminary experiment we ran ten trials with five committee
members, testing on held-out data. Figure~\ref{confidences} presents
average scores and standard deviations, varying the minimum
confidence, \textbf{m}. For a given instance, if at least \textbf{m}
committee members agreed on a tag, then the most popular tag was
applied, breaking ties in favor of the committee member that was
developed the earliest; otherwise no tag was output. The results show
that the committee approach assigns useful confidence measures to the
tags: All five committee members agreed on the tags for 45.12\% of the
instances, and 90.09\% of those tags were correct. Also, for 69.79\%
of the instances, at least four of the five committee members selected
the same tag, and this tag was correct 83.53\% of the time. We foresee
that our module for tagging dialogue acts can potentially be
integrated into a larger system so that, when Transformation-Based
Learning cannot produce a tag with high confidence, other modules may
be invoked to provide more evidence. In addition, like Boosting, the
committee method improves the overall accuracy of the system. By
selecting the most popular tag among all five committee members, the
average accuracy in tagging unseen data was 73.45\%, while using the
first committee member alone resulted in a significantly
($\mathrm{t=5.42>2.88,\ \alpha=0.01}$) lower average score of 70.79\%.

\subsection{ALTERNATIVE METHODS}

\begin{figure*}[ht]
\centerline{\psfig{figure=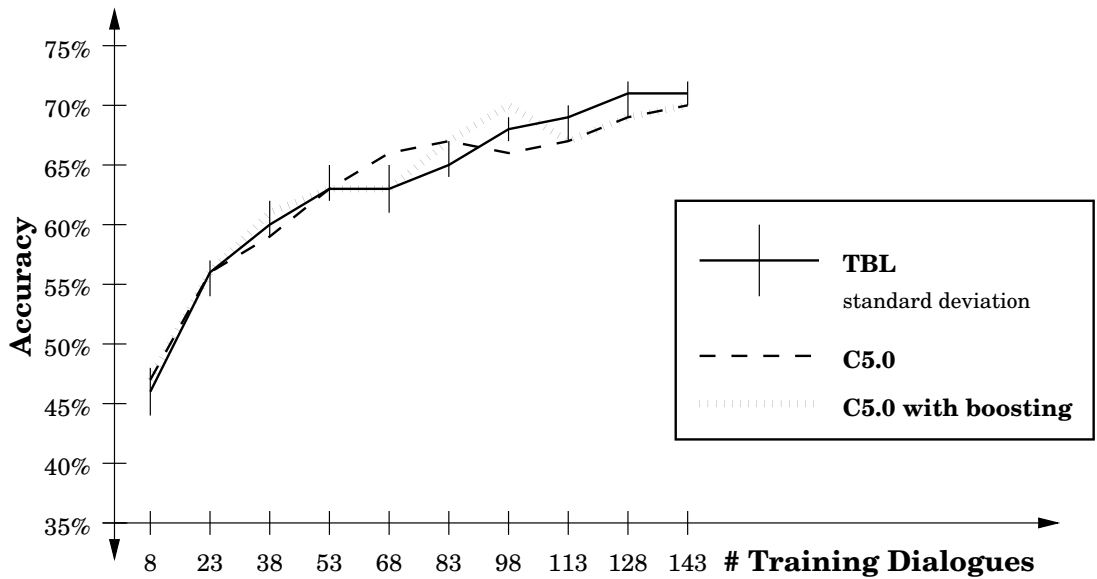}}
\caption{Training set size vs. tagging accuracy on unseen data}
\label{accuracy-graph}
\end{figure*}

Previously, the best success rate achieved on the Dialogue Act Tagging
problem was reported by Reithinger and Klesen~(1997), whose system
used a probabilistic machine learning approach based on N-Grams to
correctly label 74.7\% of the utterances in a test corpus. (See
Samuel, Carberry, and Vijay-Shanker~(1998a) for a more extensive
analysis of previous work on this task.) As a direct comparison, we
applied our system to exactly the same training and testing set. Over
five runs, the system achieved an average\footnote{The variation in
the scores is due to the random nature of the Monte Carlo method.}
accuracy of 75.12\%$\pm$1.34\%, including a high score\footnote{The
rules in Figure~\ref{ex-rules} were produced in this experiment.} of
77.44\%.

In addition, we ran a direct comparison between Transformation-Based
Learning and C5.0~(Rulequest Research, 1998), which is an
implementation of the Decision Trees method. The accuracies on
held-out data for training sets of various sizes are presented in
Figure~\ref{accuracy-graph}. For Transformation-Based Learning, we
averaged the scores of ten trials for each training set (to factor out
the random effects of the Monte Carlo method), and the standard
deviations are represented by error bars in the graph. These
experiments did not utilize the committee method, and we would expect
the scores to improve when this extension is used.

With C5.0, we wanted to use the same features that were effective for
Transformation-Based Learning, but we encountered two problems:
1)~Since C5.0 requires that each feature take exactly one value for
each instance, it is very difficult to utilize the cue patterns
feature. We decided to provide one boolean feature for each possible
cue pattern, which was set to True for instances that included that
cue pattern and False otherwise. 2)~Our Transformation-Based Learning
system utilized the system-generated tag\footnote{For
Transformation-Based Learning, the tags change as the system applies
the rules in the learned model. When a rule references a tag, it uses
the value of the tag at the point when that rule is processed.} of the
preceding instance. C5.0 cannot use this information, as it requires
that the values of all of the features are computed before training
begins.

The training times of Transformation-Based Learning and C5.0 were
relatively comparable for any number of conditions, although Boosting
sometimes resulted in a significant increase in training time. The
accuracy scores of Transformation-Based Learning and C5.0, with and
without Boosting, are not significantly different, as shown in
Figure~\ref{accuracy-graph}.

\section{DISCUSSION}

This paper has described the first investigation of
Transformation-Based Learning applied to discourse-level problems. We
extended the algorithm to address two limitations of
Transformation-Based Learning: 1)~We developed a Monte Carlo version
of Transformation-Based Learning, and our experiments suggest that
this improvement dramatically increases the efficiency of the method
without compromising accuracy. This revision enables
Transformation-Based Learning to work effectively on a wider variety
of tasks, including tasks where the relevant conditions and condition
combinations are not known in advance as well as tasks where there are
a large number of relevant conditions and condition combinations. This
improvement also decreases the labor demands on the human developer,
who no longer needs to construct a minimal set of rule templates. It
is sufficient to list all of the conditions that might be relevant and
allow the system to consider all possible combinations of those
conditions. 2)~We devised a committee strategy for computing
confidence measures to represent the reliability of tags. In our
experiments, this committee method improved the overall tagging
accuracy significantly. It also produced useful confidence measures;
nearly half of the tags were assigned high confidence, and of these,
90\% were correct.

For the Dialogue Act Tagging task, our modified version of
Transformation-Based Learning has achieved an accuracy rate that is
comparable to any previously reported system. In addition,
Transformation-Based Learning has a number of features that make it
particularly appealing for the Dialogue Act Tagging task:

\begin{enumerate}
\item Transformation-Based Learning's learned model consists of a
relatively short sequence of intuitive rules, stressing relevant
features and highlighting important relationships between features and
tags~(Brill, 1995a). Thus, Transformation-Based Learning's learned
model offers insights into a {\em theory} to explain the training
data. This is especially useful in Dialogue Act Tagging, which
currently lacks a systematic theory.
\item With its iterative training algorithm, when developing a new
rule, Transformation-Based Learning can consider tags that have been
produced by previous rules~(Ramshaw and Marcus, 1994). Since the
dialogue act of an utterance is affected by the surrounding dialogue
acts, this leveraged learning approach can directly integrate the
relevant contextual information into the rules. In addition,
Transformation-Based Learning can accommodate the focus shifts that
frequently occur in discourse by utilizing features that consider tags
of varying distances.
\item Our Transformation-Based Learning system is very flexible with
respect to the types of features it can utilize. For example, it can
learn set-valued features, such as cue patterns. Additionally, because
of the Monte Carlo improvement, our system can handle a very large
number of features.
\item For the Dialogue Act Tagging task, people still don't know what
features are relevant, so it is very difficult to construct an
appropriate set of rule templates. Fortunately, Transformation-Based
Learning is capable of discarding irrelevant rules, as Ramshaw and
Marcus~(1994) showed experimentally, so it is not necessary that {\em
all} of the given rule templates be useful.
\item Ramshaw and Marcus's~(1994) experiments suggest that
Transformation-Based Learning tends to be resistant to the
overfitting\footnote{Other machine learning algorithms may overfit to
the training data and then have difficulty generalizing to new data.}
problem. This can be explained by observing how the rule sequence
produced by Transformation-Based Learning progresses from general
rules to specific rules. The early rules in the sequence are based on
many examples in the training corpus, and so they are likely to
generalize effectively to new data. Later in the sequence, the rules
don't receive as much support from the training data, and their
applicability conditions tend to be very specific, so they have little
or no effect on new data. Thus, resistance to overfitting is an
emergent property of the Transformation-Based Learning algorithm.
\end{enumerate}

For the future, we intend to investigate a wider variety of features
and explore different methods for collecting cue patterns to increase
our system's accuracy scores further. Although we compared
Transformation-Based Learning with a few very different machine
learning algorithms, we still hope to examine other methods, such as
Naive Bayes. In addition, we plan to run our experiments with
different corpora to confirm that the encouraging results of our
extensions to Transformation-Based Learning can be generalized to
different data, languages, domains, and tasks. We would also like to
extend our system so that it may learn from untagged data, as there is
still very little tagged data available in discourse. Brill developed
an unsupervised version of Transformation-Based Learning for
Part-of-Speech Tagging~(Brill, 1995b), but this algorithm must be
initialized with instances that can be tagged unambiguously (such as
``the'', which is always a determiner), and in Dialogue Act Tagging
there are very few unambiguous examples. We intend to investigate the
following weakly-supervised approach: First, the system will be
trained on a small set of tagged data to produce a number of different
committee members. Then given untagged data, it will derive tags with
confidence measures. Those tags that receive very high confidence can
be used as unambiguous examples to drive the unsupervised version of
Transformation-Based Learning.

\subsubsection*{Acknowledgments}
 
We wish to thank the members of the \textsc{VerbMobil} research group
at DFKI in Germany, particularly Norbert Reithinger, Jan
Alexandersson, and Elisabeth Maier, for providing the first author
with the opportunity to work with them and generously granting him
access to the \textsc{VerbMobil} corpora. This work was partially
supported by the NSF Grant \#GER-9354869.

\subsubsection*{References} 

Brill, Eric (1995a). Transformation-Based Error-Driven Learning and
Natural Language Processing: A Case Study in Part-of-Speech Tagging.
{\it Computational Linguistics} {\bf 21}(4):543-566.

Brill, Eric (1995b). Unsupervised Learning of Disambiguation Rules for
Part of Speech Tagging. In {\it Proceedings of the Very Large Corpora
Workshop}.

Brill, Eric and Mooney, Raymond~J. (1997). An Overview of Empirical
Natural Language Processing. {\it AI Magazine} {\bf 18}(4):13-24.

Dagan, Ido and Engelson, Sean~P. (1995). Committee-Based Sampling for
Training Probabilistic Classifiers. In {\it Proceedings of the Twelfth
International Conference on Machine Learning}.

Freund, Yoav and Schapire, Robert~E. (1996). Experiments with a New Boosting
Algorithm. In {\it Proceedings of the Thirteenth International
Conference on Machine Learning}.

Hirschberg, Julia and Litman, Diane (1993). Empirical Studies on the
Disambiguation of Cue Phrases. {\it Computational Linguistics} {\bf
19}(3):501-530.

Knott, Alistair (1996). A Data-Driven Methodology for Motivating a Set
of Coherence Relations. {\it Ph.D. Thesis}. The University of
Edinburgh.

Ramshaw, Lance~A. and Marcus, Mitchell~P. (1994). Exploring the Statistical
Derivation of Transformation Rule Sequences for Part-of-Speech
Tagging. In {\it Proceedings of the 32nd Annual Meeting of the ACL}.
 
Reithinger, Norbert and Klesen, Martin (1997). Dialogue Act Classification Using
Language Models. In {\it Proceedings of EuroSpeech-97}.

Rulequest Research. (1998). Data Mining Tools see5 and c5.0.
[http://www.rulequest.com/see5-info.html].

Samuel, Ken, Carberry, Sandra, and Vijay-Shanker, K. (1998a). Computing
Dialogue Acts from Features with Transformation-Based Learning. In
{\em Applying Machine Learning to Discourse Processing: Papers from
the 1998 AAAI Spring Symposium}.

Samuel, Ken, Carberry, Sandra, and Vijay-Shanker, K. (1998b). Dialogue Act
Tagging with Transformation-Based Learning. In {\em Proceedings of
COLING-ACL}.

\end{document}